\documentclass[conference]{IEEEtran}
\usepackage{subcaption}
\IEEEoverridecommandlockouts
\usepackage{cite}
\usepackage{amsmath,amssymb,amsfonts}
\usepackage{algorithmic}
\usepackage{graphicx}
\usepackage{textcomp}
\usepackage{xcolor}
\usepackage[utf8]{inputenc}
\def\BibTeX{{\rm B\kern-.05em{\sc i\kern-.025em b}\kern-.08em
    T\kern-.1667em\lower.7ex\hbox{E}\kern-.125emX}}
\begin{document}

\title{DPA-Net: A Dual-Path Attention Neural Network for Inferring Glycemic Control Metrics from Self-Monitored Blood Glucose Data\\

\thanks{* Corresponding author. Contacting email: hcf7fd@virginia.edu}
}

\author{\IEEEauthorblockN{Canyu Lei $ ^1$, Benjamin Lobo $ ^2$, Jianxin Xie $ ^{2*}$}
\IEEEauthorblockA{1 Binghamton University, Department of Computer Science \\
2 University of Virginia, School of Data Science
}
}

\maketitle

\begin{abstract}



Continuous glucose monitoring (CGM) provides dense and dynamic glucose profiles that enable reliable estimation of Ambulatory Glucose Profile (AGP) metrics, such as Time in Range (TIR), Time Below Range (TBR), and Time Above Range (TAR). However, the high cost and limited accessibility of CGM restrict its widespread adoption, particularly in low- and middle-income regions. In contrast, self-monitoring of blood glucose (SMBG) is inexpensive and widely available, but yields sparse and irregular data that are challenging to translate into clinically meaningful glycemic metrics.
In this work, we propose a Dual-Path Attention Neural Network (DPA-Net), to estimate AGP metrics directly from SMBG data. DPA-Net integrates two complementary paths: (1) a spatial–channel attention path that reconstructs a CGM-like trajectory from sparse SMBG observations, and (2) a multi-scale ResNet path that directly predicts AGP metrics. An alignment mechanism between the two paths is introduced to reduce bias and mitigate overfitting. In addition, we develop an active point selector to identify realistic and informative SMBG sampling points that reflect patient behavioral patterns.
Experimental results on a large, real-world dataset demonstrate that DPA-Net achieves robust accuracy with low errors, while reducing systematic bias. To the best of our knowledge, this is the first supervised machine learning framework for estimating AGP metrics from SMBG data, offering a practical and clinically relevant decision-support tool in settings where CGM is not accessible.

\end{abstract}



\section{Introduction}

With the steadily increasing prevalence of diabetes, it has become one of the most common and challenging chronic diseases worldwide, imposing a substantial burden on public health\cite{hossain2024diabetes}. Diabetes is a major cause of mortality and morbidity and is closely associated with cardiovascular disease, renal failure, and other complications\cite{khunti2023diabetes}. For individuals with diabetes, timely monitoring and effective control of blood glucose are critical to reducing complications and improving health outcomes\cite{kirk2010self}.

Time in range (TIR) has been widely used as a key indicator of glycemic control, and the monitoring goal for individuals with diabetes in daily life is to remain within this range as much as possible \cite{goshrani2025time}. A common approach is self-monitoring of blood glucose (SMBG), which typically requires pricking a finger to obtain a small blood sample \cite{benjamin2002self}. As a result, patients usually perform only a limited number of measurements per day; the data are not only sparse but also highly susceptible to various interfering factors. Therefore, clinicians face considerable challenges in using such data to guide patients in managing their diabetes \cite{erbach2016interferences}.

The Ambulatory Glucose Profile (AGP) is a “standardized, practical one-page report” originally designed to provide continuous glucose monitoring (CGM) users with a clear visualization of their glycemic control \cite{czupryniak2022ambulatory}. The report was later adapted for SMBG users, deriving measures of glycemic variability from limited SMBG data. However, compared with metrics derived from CGM, the SMBG version often exhibits systematic bias \cite{avari2020differences}. For example, in adults with well-controlled type 1 diabetes (T1D), studies have shown that SMBG-based assessments tend to systematically underestimate TIR, overestimate TAR and overestimate TBR, \cite{aleppo2017replace}. These biases are observed across different analysis windows (two weeks, one month, or one year) and become more pronounced as the window shortens. Since the SMBG version of AGP relies on a two-week window, clinicians are inevitably required to make decisions based on these biased metrics. 

By contrast, CGM provides continuous and high-frequency glucose measurements, enabling more accurate and unbiased estimation of AGP metrics. As a result, CGM-based monitoring has been shown to be more effective in optimizing glycemic management than SMBG, reducing hypoglycemic events, and improving patient satisfaction \cite{janapala2019continuous, bergenstal2022randomized}.


Although CGM substantially guarantees data density and clinical utility for estimating the status of blood glucose, its widespread adoption is constrained by multiple factors and remains largely restricted to high-income countries\cite{huang2010cost}. In many low- and middle-income countries (LMICs), SMBG continues to serve as the predominant method of glucose monitoring \cite{ewen2025availability}. Moreover, CGM may be unsuitable for certain populations, such as adolescents and young adults with T1D, due to challenges related to cost, wearability, and adherence \cite{datye2021advances}.


The objective of this study is to utilize sparse SMBG data to predict 3-level AGP glycemic metrics, including TIR, TBR, and TAR. These metrics reflect the quality of glycemic control, capturing the severity of hypo- and hyperglycemia, and are widely recognized as clinically meaningful indicators of diabetes management. Our goal is to achieve accurate estimation of AGP metrics from SMBG data that is comparable to those obtained from CGM.


To accomplish this, we propose a Dual Path Spatial-Channel Self-Attention Network (DPA-Net), designed to maintain high predictive accuracy while substantially lowering the frequency of blood glucose measurements needed. The model consists of two complementary paths: one path reconstructs the CGM curve to capture latent dynamics from sparse SMBG data, and the other directly predicts glycemic control metrics—TAR, TIR, and TBR—thereby ensuring both temporal consistency and clinical utility. The main contributions and significance of this work are as follows:
\begin{itemize}
   
   \item We propose DPA-Net, a tailored architecture for estimating AGP metrics from SMBG. It incorporates a spatial–channel attention path to reconstruct potential CGM trajectories and a Multi-scale ResNet path to capture multi-scale features from SMBG inputs for direct glycemic prediction. By enforcing consistency between the two paths, the model effectively mitigates overfitting and improves predictive accuracy.
   \item To reduce bias and better reflect real-world conditions—where patients prick fingers according to patterns rather than randomly—we propose an active point selector, which learns SMBG behavior patterns to identify the most likely measurement times among the history CGM, enabling supervised training on more realistic data and improving robustness and clinical relevance.
   \item The approach offers a cost-conscious alternative for settings where CGM is not yet widely accessible, with great potential in clinical utility and public-health impact in LMICs.
\end{itemize}
To the best of our knowledge, no prior studies have directly attempted to estimate AGP glycemic metrics from SMBG data. This underscores the novelty of our work, while also limiting the availability of established benchmarks for direct comparison.

\section{Related Work}

Although SMBG has limitations in accuracy and data density\cite{schnell2015clinical}, it served for a long time as the predominant method of glucose monitoring before CGM was introduced\cite{hirsch2008self, briggs2004self}. To this end, early studies proposed a variety of traditional SMBG-based indices to characterize glucose levels and variability from limited sampling data.

Specifically, one of the earliest indices developed to quantify glycemic variability was the Mean Amplitude of Glycemic Excursions (MAGE), proposed by Service et al. in the 1970s\cite{service1970mean}, which became a representative measure of glucose fluctuations. Subsequently, mean glucose and standard deviation (Mean and SD) were widely applied in landmark clinical studies such as the Diabetes Control and Complications Trial(DCCT)\cite{effect1993diabetes} and Kumamoto trials\cite{ohkubo1995intensive} to systematically evaluate glycemic control. Later, Wojcicki introduced the J-index, which combines both glucose levels and variability into a single comprehensive metric\cite{wojcicki1995j}, while Kovatchev et al. proposed the Low and High Blood Glucose Indices (LBGI/HBGI) to characterize the risks associated with extreme glucose values\cite{kovatchev1997symmetrization}. In addition, the coefficient of variation (CV\%) has been increasingly adopted as a standardized measure to compare the magnitude of glucose variability across different patients and populations\cite{monnier2008glycemic}. Furthermore, before the widespread adoption of CGM, clinical practice guidelines also incorporated several of the above-mentioned metrics, such as mean glucose, SD, and MAGE, as surrogate indicators of glycemic control\cite{american2005standards,american2006standards}, despite their limited ability to capture glucose dynamics at the resolution provided by AGP reports.

With the advancement of artificial intelligence, an increasing number of machine learning methods have been applied to the management and prediction of diabetes\cite{mujumdar2019diabetes, alam2024machine,nomura2021artificial, afsaneh2022recent}, including several explorations based on SMBG data. Several studies have developed predictive models based on SMBG data to identify the risk of hypoglycemia or hyperglycemia. For example, Sudharsan et al. developed machine learning models to predict hypoglycemic events in patients with type 2 diabetes using SMBG data\cite{sudharsan2014hypoglycemia}. In addition, Oviedo et al.  trained individualized models to predict postprandial hypoglycemia based on capillary SMBG data from patients with diabetes using various machine learning algorithms, achieving more accurate identification of postprandial hypoglycemia risk\cite{oviedo2019minimizing}. These studies demonstrate that, even with sparse data, SMBG can still provide value for individualized risk prediction; however, they remain limited to single-event prediction and lack modeling of global indicators of glycemic control.

Beyond event-level prediction, another line of research has focused on using SMBG data for pattern recognition and variability assessment, aiming to provide a more comprehensive characterization of glycemic control. Ljiljana et al. used routine primary care data including 6–8 random SMBG measurements and found that patients with diabetes exhibited high glucose variability in both fasting and postprandial states\cite{ljiljana2020use}. Li et al. applied machine learning methods to model limited SMBG records in order to predict HbA1c levels in patients with type 2 diabetes, exploring how to infer long-term glycemic control without relying on extended follow-up\cite{li2024minimized}. Faruqui et al. developed a deep learning model to forecast near-future glucose levels in patients with diabetes using routine data, including SMBG records\cite{faruqui2019development}. Woldaregay et al. applied machine learning techniques to type 1 diabetes data to classify blood glucose patterns and detect anomalies, demonstrating the potential of SMBG-based modeling for identifying irregular glycemic behaviors\cite{woldaregay2019data}. These studies demonstrate that machine learning can provide valuable assessments of glucose patterns and variability under sparse SMBG conditions. 
However, these approaches have largely been limited to short-term or small-scale datasets and have not systematically investigated AGP metrics or demonstrated comprehensive assessments of glycemic control status in individuals with diabetes.

\section{Methodology}

\subsection{Problem Statement}
\begin{figure}[htbp]
\centering
\includegraphics[width=0.5\linewidth]{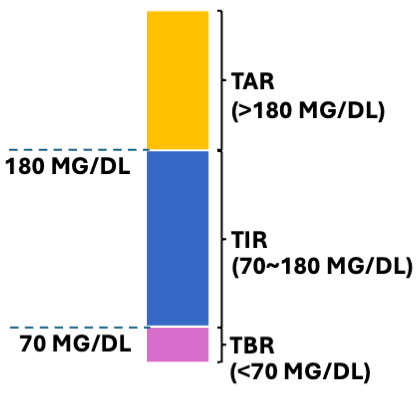}
\caption{AGP times-in-range “thermometer” (TAR/TIR/TBR).}
\label{fig:tr_description}
\end{figure}

Continuous Glucose Monitoring (CGM) provides dense glucose readings and is widely regarded as the gold standard for assessing glycemic control\cite{cappon2019contimuous}. Over a two-week monitoring window, the glycemic variability metrics from the Ambulatory Glucose Profile (AGP) report, including Time in Range (TIR), Time Above Range (TAR), and Time Below Range (TBR), can be computed directly and reliably from CGM traces, offering an accurate summary of a patient’s glycemic state. Specifically, TIR was defined as the proportion of CGM readings within the range 70–180 mg/dL, TAR as the proportion above $\tau_\text{high} = 180$ mg/dL, and TBR as the proportion below $\tau_\text{low} = 70$  mg/dL. Figure~\ref{fig:tr_description} illustrates these label definitions. 

Let $D$ denote the number of observation days ($D = 14$ in this study), and let each day be discretized into $T$ time bins of equal time length($T = 288$ for a 5-minute interval). AGP metrics are then computed by counting the fraction of CGM samples that fall within each glycemic region (in-range, above-range, below-range) over the two-week window: 

\begin{align}
\label{Eq:cgm_to_tr}
\mathrm{TIR}_\rho &=\frac{1}{D T} \sum_{d=1}^D \sum_{t=1}^T \mathbf{1}\left(\tau_{\text {L }} \leq g_{d, t}^{(\rho)} \leq \tau_{\text {high }}\right)  \nonumber\\
\mathrm{TAR}_\rho &=\frac{1}{D T} \sum_{d=1}^D \sum_{t=1}^T \mathbf{1}\left(g_{d, t}^{(\rho)}>\tau_{\text {high }}\right) \nonumber\\
\mathrm{TBR}_\rho &=\frac{1}{D T} \sum_{d=1}^D \sum_{t=1}^T \mathbf{1}\left(g_{d, t}^{(\rho)}<\tau_{\text {low }}\right)
\end{align}

where $g_{d,t}$ denote the CGM glucose value at day $d$ and time instance $t$ for sample $\rho$. By construction, $\text{TAR}_\rho + \text{TIR}_\rho + \text{TBR}_\rho =1 $.

 Although CGM provides accurate and convenient computation of AGP metrics, its use remains limited in many settings due to cost, device availability, and adherence challenges\cite{barchiesi2025continuous}. In contrast, SMBG is more widely accessible but produces only sparse and irregularly sampled measurements. To enable AGP estimation in contexts where CGM is not feasible, we represent the two-week SMBG observation window in a structured form as follows.

 For each sample instance, two consecutive weeks of SMGB data were collected. 
 The two-week observation window for an SMBG sample is represented as 
\begin{equation}
\label{Eq:smbg_input}
    M_s = \{s_{d,t}\mid d=1,\cdots,D; t = 1,\cdots,T\} \in \mathbb{R}^{D\times T}
\end{equation}
where $s_{d,t}$ denotes the glucose value on day $D$ at time instance $t$, and $s_{d,t}=0$ if no SMBG measurement is observed at $(d,t)$.



Suppose there are $N$ samples indexed by $\rho \in\{1,\cdots, N\}$. Each sample can then be denoted as a pair $(M_{s,\rho}, \text{TR}_\rho)$, where $M_{s,\rho}$ is the SMBG observation matrix for sample $\rho$, and $\text{TR}_p$ denotes the corresponding set of glycemic variability metrics from AGP report, i.e., $\text{TR}_\rho = \{\text{TAR}, \text{TIR}, \text{TBR}\}_\rho$. The objective of this task is learning a function $f_\theta$ that maps the SMBG data $M_{s,\rho}$ to the target glycemic metrix $\text{TR}_\rho$ for two weeks that accurately reflects glycemic state, i.e., 
\begin{equation}
   f_\theta :  M_{s,\rho}\longrightarrow \widehat{TR}_{\rho}
\end{equation}
where $f_\theta$ is parameterized by the proposed dual-path neural attention network.



\subsection{Dual-Path Model}
\begin{figure*}
\centering
\includegraphics[width=0.8\linewidth]{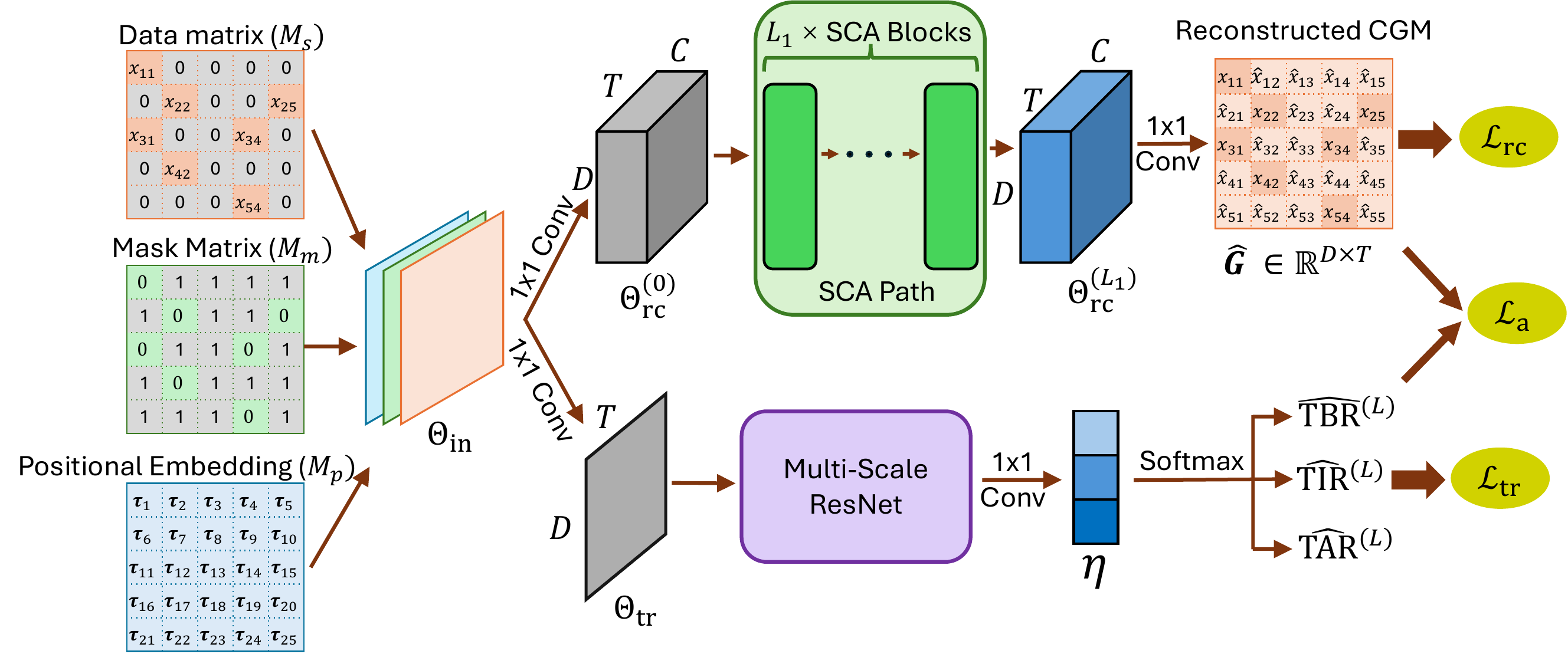}
\caption{Overall architecture of the proposed Dual-Path Attention Network (DPA-Net). 
The input $\Theta_{\mathrm{in}}$ combines the data matrix ($M_s$), mask matrix ($M_m$), 
and positional embedding ($M_p$). The upper SCA Path reconstructs dense CGM from sparse SMBG, while the lower multi-scale ResNet path directly predicts long-term glycemic range metrics (TIR, TAR, TBR). 
Three complementary loss functions ($L_{\mathrm{rc}}, L_a, L_{\mathrm{tr}}$) jointly supervise training.}
\label{fig:overall_dualpath}
\end{figure*}

In this study, we propose a Dual-Path Attention Network (DPA-Net) to predict long-term glycemic control metrics—TAR, TIR, TBR—over a $D$-day observation period using only sparse Self-Monitoring of Blood Glucose (SMBG) data. The model pipeline is illustrated in Fig. \ref{fig:overall_dualpath}.

The input consists of three components: 
\begin{enumerate}
    \item \textbf{SMBG observation matrix $M_{s,\rho}$:} represents the two-week SMBG observation window, as defined in Eq. \ref{Eq:smbg_input};
    \item \textbf{Missing-value mask matrix $M_{m,\rho}$:} a binary matrix that explicitly encodes the sampling pattern of the SMBG data for sample $\rho$. Since SMBG readings are sparse and irregularly distributed across the $D\times T$ observation grid, it is important for the model to distinguish between (i) bins where no glucose measurement was taken and (ii) bins where the glucose level is legitimately zero (which never occurs physiologically, but would otherwise be numerically confounded if we simply inserted zeros for missing data). The mask matrix therefore provides the network with information about where values are observed versus missing, preventing misinterpretation of unobserved entries. Formally,
    \begin{equation}
    M_m = \{m_{d,t}\mid d=1,\cdots,D; t = 1,\cdots,T\} \in \mathbb{R}^{D\times T} 
    \end{equation}
    where 
    \begin{equation}
    m_{d,t} =
    \begin{cases}
    1, & \text{if the BG is missing at time } t \\
    0, & \text{otherwise}
    \end{cases}
    \end{equation}

    This mask is used as an additional input channel to the model, allowing the network to learn from both the glucose values that are present and the pattern of missingness itself, which carries clinically relevant information about patient sampling behavior.

    \item \textbf{Positional encoding matrix $M_{p,\rho}$:} an embedding matrix that encodes temporal information so that the model is aware of the natural order of the glucose series after reshaping into the $D\times T$ grid data. 

    Since the 2D matrix structure does not inherently preserve the original temporal order, we incorporate a 2D multi-frequency sinusoidal positional encoding, which enable the model to simultaneously capture intra-day temporal order and inter-day periodicity in the 2D representation.  

 
    We incorporate a two-dimensional positional encoding. Following the standard sinusoidal scheme\cite{vaswani2017attention}, encodings are generated independently for the inter-day and intra-day dimensions and subsequently combined. Let $p$ denote the embedding index and $P$ represent the total encoding dimension. For the day dimension, each day index $i\in\{0,\cdots,D-1\}$ is encoded as:

\begin{equation}
\begin{split}
\mathrm{PE}_{\mathrm{day}}[i, 2p]   &= \sin\left(\frac{i}{10000^{2p/P}}\right), \\
\mathrm{PE}_{\mathrm{day}}[i, 2p+1] &= \cos\left(\frac{i}{10000^{2p/P}}\right)
\end{split}
\end{equation}
where $i = 0, 1, \dots, D-1$.

Similarly, for the time-of-day dimension, each time index $j\in\{0,\cdots,T-1\}$ is encoded as
\begin{equation}
\begin{split}
    \mathrm{PE}_{\mathrm{time}}[j, 2p]   = \sin\left(\frac{j}{10000^{2p/P}}\right), \\
\mathrm{PE}_{\mathrm{time}}[j, 2p+1] = \cos\left(\frac{j}{10000^{2p/P}}\right)
\end{split}
\end{equation}
where $j = 0, 1, \dots, T_D-1$.



Then we combined the two encodings $\mathrm{PE}_{\mathrm{day}}$ and $\mathrm{PE}_{\mathrm{time}}$ by element-wise summation, producing a positional encoding at each day–time location. To match the input channel format, we then collapse the encoding dimension into a single channel by summing across it, yielding a matrix $M_p \in \mathbb{R}^{D \times T}$.

\end{enumerate}

As illustrated in Figure~\ref{fig:overall_dualpath}, the composite input feature of DPA-Net, denoted as $\Theta_{\mathrm{in}}$, integrates the glucose value matrix $M_s$, the missing-value mask $M_m$, and the positional encoding $M_p$. Formally, 
\begin{equation}
    \Theta_{\mathrm{in}} = \mathrm{Concat}\!\left(M_s,\, M_m,\, M_p\right) \in \mathbb{R}^{3 \times D \times T_D}
\end{equation}

and is subsequently processed through two complementary paths. The upper path aims to approximate the underlying CGM trajectory over the two-week observation window by reconstructing a dense glucose time series from the sparse SMBG inputs. From this reconstructed trajectory, surrogate AGP metrics $\widehat{\text{TR}}_\rho^{(U)} = \{\widehat{\text{TAR}}, \widehat{\text{TIR}}, \widehat{\text{TBR}}\}_\rho^{(U)}$ can be computed in the same manner as with true CGM data according to Eq.\ref{Eq:cgm_to_tr}. In contrast, the lower path avoids trajectory reconstruction and directly maps the SMBG representation to predicted glycemic metrics $\widehat{\text{TR}}_\rho^{(L)}$. By aligning the outcomes of the two paths during training, the model learns both to recover physiologically plausible glucose dynamics and to generate robust estimates of glycemic metrics.The reconstruction path is critical to this process: rather than fitting directly to the target metrics alone, the model is encouraged to approximate the underlying CGM trajectory from sparse SMBG inputs. This additional constraint reduces the risk of overfitting by embedding physiological structure into the learning process, as the reconstructed trajectory must remain consistent with realistic glucose fluctuations across the two-week period.





\subsection{Spatial-Channel Attention Path for Continuous Blood Glucose Reconstruction}

The upper path of the proposed model, namely the Spatial-Channel Attention (SCA) Path, is designed to reconstruct the continuous CGM trajectory from sparse SMBG inputs. This process enables the network to recover fine-grained glucose dynamics that are otherwise unobserved in SMBG data.

The composite input $\Theta_{\mathrm{in}}$ is first projected through a convolution layer. This operation maps the three-channel representation into a higher-dimensional feature space and produces an input feature map for upper path: $\Theta_{\mathrm{rc}}^{(0)} \in \mathbb{R}^{C \times D \times T}$ where $C$ denotes the expanded channel dimension. This feature map serves as the basis for subsequent attention modules that jointly capture intra-day and inter-day temporal dependencies and channel-wise correlations.

\begin{figure}[htbp]
\centering
\includegraphics[width=0.9\linewidth]{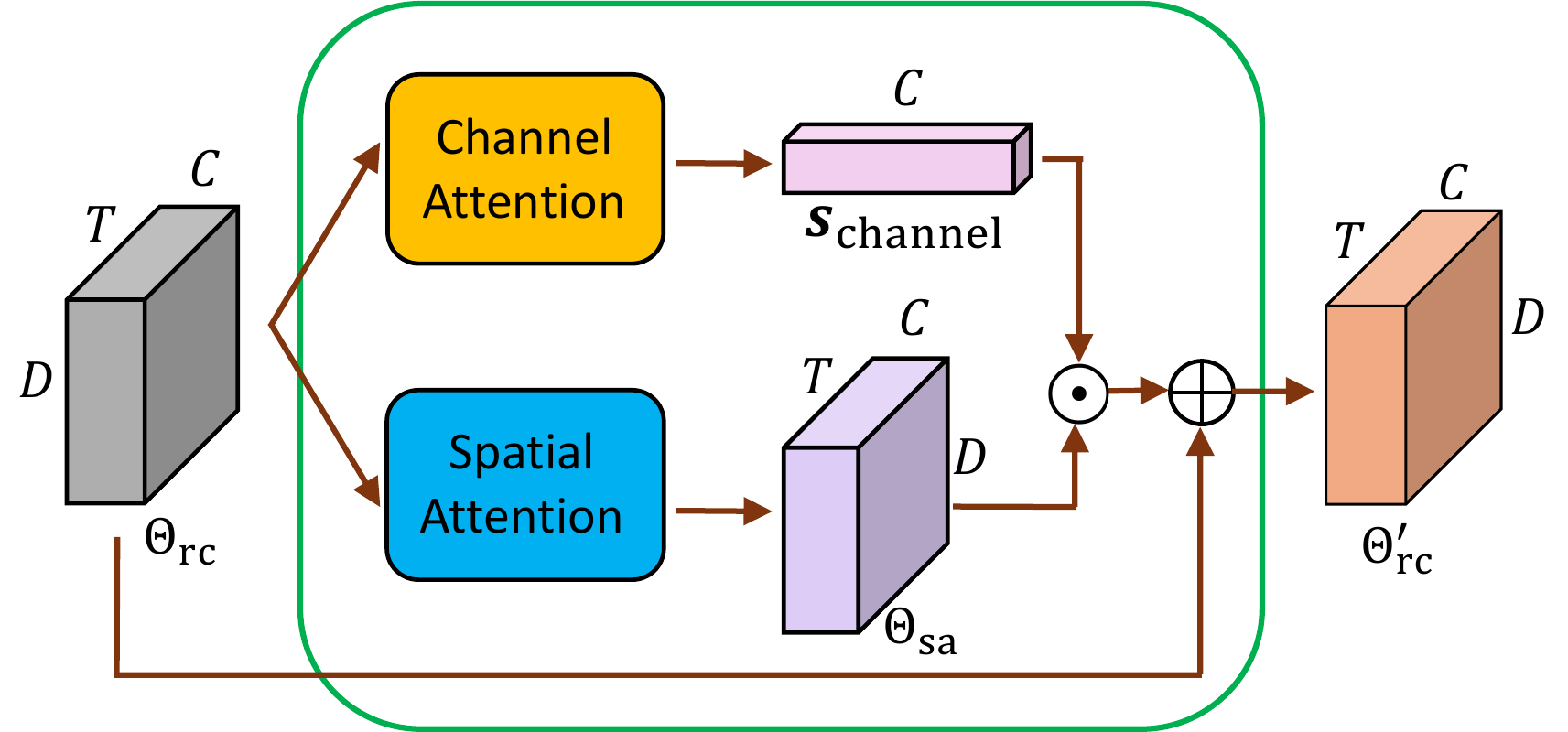}
\caption{Network structure for Spatial-Channel Attention (SCA).}
\label{fig:sca}
\end{figure}

The upper path is composed of $L$ Spatial-Channel Attention Blocks arranged in series. Each block, as illustrated in the Fig. \ref{fig:sca}, consists of two key components: a Channel Attention Block and a Spatial Attention Block:

\begin{figure}[htbp]
\centering
\includegraphics[width=0.95\linewidth]{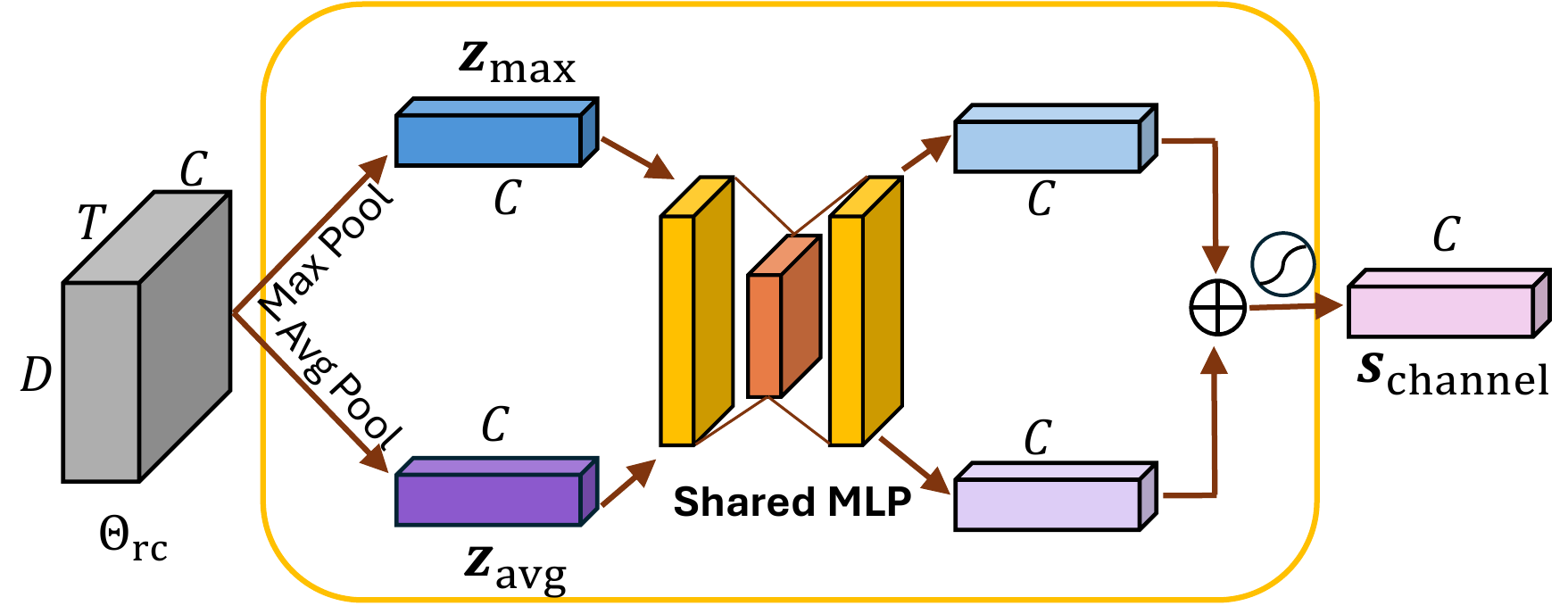}
\caption{Channel attention block.}
\label{fig:channel_attention}
\end{figure}

\begin{enumerate}
    \item \textbf{Channel attention block:} Channel attention Block models the relative importance of different feature channels and adaptively enhances the most informative ones through a dynamic weighting mechanism\cite{hu2018squeeze}. In each Spatial-Channel Attention Block, we adopt a channel attention\cite{woo2018cbam}, which combines global average pooling\cite{lin2013network} and max pooling\cite{woo2018cbam} descriptors to generate channel-wise weights for re-scaling the feature responses. The structure of the channel attention is illustrated in Fig. \ref{fig:channel_attention}.

    Specifically, given a feature map
    $\Theta_{\text{rc}} \in \mathbb{R}^{C \times D \times T}$, spatial information is first aggregated through average and max pooling, producing two-channel descriptors that capture the global context of each channel:
    
    
    
    \begin{align}
        \boldsymbol{z}_{\mathrm{avg}} &= [z_{\mathrm{avg}}^1,z_{\mathrm{avg}}^2,\dots,z_{\mathrm{avg}}^C] \in \mathbb{R}^C \nonumber\\
        \boldsymbol{z}_{\mathrm{max}} &= [z_{\mathrm{max}}^1,z_{\mathrm{max}}^2,\dots,z_{\mathrm{max}}^C] \in \mathbb{R}^C
    \end{align}

    The aggregated descriptors are passed through a shared two-layer MLP, and their outputs are combined by element-wise summation followed by a sigmoid activation, yielding the channel weight vector $s_{\mathrm{channel}}$.
    
    \begin{equation}
        \boldsymbol{s}_{\mathrm{channel}} = \sigma \left( W_2 \, \delta \left( W_1 z_{\mathrm{avg}} \right) 
        + W_2 \, \delta \left( W_1 z_{\mathrm{max}} \right) \right),
    \end{equation}
    
    where $W_1 \in \mathbb{R}^{\frac{C}{r_c} \times C}$, 
    $W_2 \in \mathbb{R}^{C \times \frac{C}{r_c}}$, $r_c$ is the reduction ratio for channel attention, a hyperparameter that controls the dimensionality reduction in the bottleneck. $\delta(\cdot)$ denotes the ReLU activation, and $\sigma(\cdot)$ denotes the sigmoid function. The channel attention block adaptively provideds the reweight of each channel by modeling global SMBG information, enabling the network to emphasize informative features while suppressing less relevant ones. This mechanism enhances representational capacity without introducing significant computational overhead.

    \item \textbf{Spatial attention block:} Spatial Attention Block captures temporal dependencies within a day as well as periodic patterns across days, ensuring that both short-term fluctuations and long-term trends are effectively represented. The spatial self-attention aims to capture long-range dependencies across spatial positions and enhance the contextual representation of each location\cite{zhang2019self}. The structure of the spatial self-attention module is illustrated in Fig. \ref{fig:sca}.

    Similar to channel attention, the input is the feature map $\Theta_{\text{rc}} \in \mathbb{R}^{C \times D \times T}$.
    First, we obtain the query $Q$, key $K$, and value $V$ matrices by applying three independent $1\times 1$ convolutions\cite{vaswani2017attention}:
    
    \begin{equation}
        \begin{split}
            Q = W_Q * \Theta_{\mathrm{rc}} \in \mathbb{R}^{C'\times D\times T}, \\
            K = W_K * \Theta_{\mathrm{rc}} \in \mathbb{R}^{C'\times D\times T}, \\
            V = W_V * \Theta_{\mathrm{rc}} \in \mathbb{R}^{C\times D\times T}, 
        \end{split}
    \end{equation}
    
    where $W_Q, W_K \in \mathbb{R}^{\frac{C}{r_s} \times C}$ are learnable projection matrices with reduction ratio $r_s$, $C' = \tfrac{C}{r_s}$), and $W_V \in \mathbb{R}^{C \times C}$ preserves the original channel dimension.
    Here the $1\times 1$ convolution performs channel-wise projection while maintaining the same spatial dimension $(D, T)$.
    
    To enables the computation of pairwise dependencies between all spatial positions through subsequent attention operations, we compress the spatial dimensions $(D, T)$ into a single dimension after obtain the Query, Key, and Value matrices\cite{zhang2019self}.
    
    \begin{equation}
    \begin{split}
        Q' \in \mathbb{R}^{(D \cdot T) \times C'}, \\
        K' \in \mathbb{R}^{C' \times (D \cdot T)}, \\
        V' \in \mathbb{R}^{C \times (D \cdot T)}
    \end{split}
    \end{equation}
    
    where $C'$ is the reduced channel dimension, 
    and $C$ is the original channel dimension. 
    
    The position-to-position attention map is computed as
    \begin{equation}
        A = \mathrm{Softmax}\!\left(Q' K'\right) 
    \in \mathbb{R}^{(D \cdot T) \times (D \cdot T)}
    \end{equation}
    
    We then aggregate the Value features with the attention weights:
    \begin{equation}
        O' = V' A \;\in\; \mathbb{R}^{C \times (D \cdot T)}
    \end{equation}
    
    The aggregated output is reshaped back to the original spatial dimensions:
    \begin{equation}
        O \in \mathbb{R}^{C \times D \times T}
    \end{equation}
    
    Finally, a residual connection with a learnable scalar parameter is applied to stabilize training and control the contribution of attention:
    \begin{equation}
        \Theta_{\mathrm{sa}} = \gamma_1 \cdot O + \Theta_{\mathrm{rc}},
    \end{equation}
    
    where $\gamma$ is a trainable parameter.
\end{enumerate}

\begin{figure}[htbp]
\centering
\includegraphics[width=0.85\linewidth]{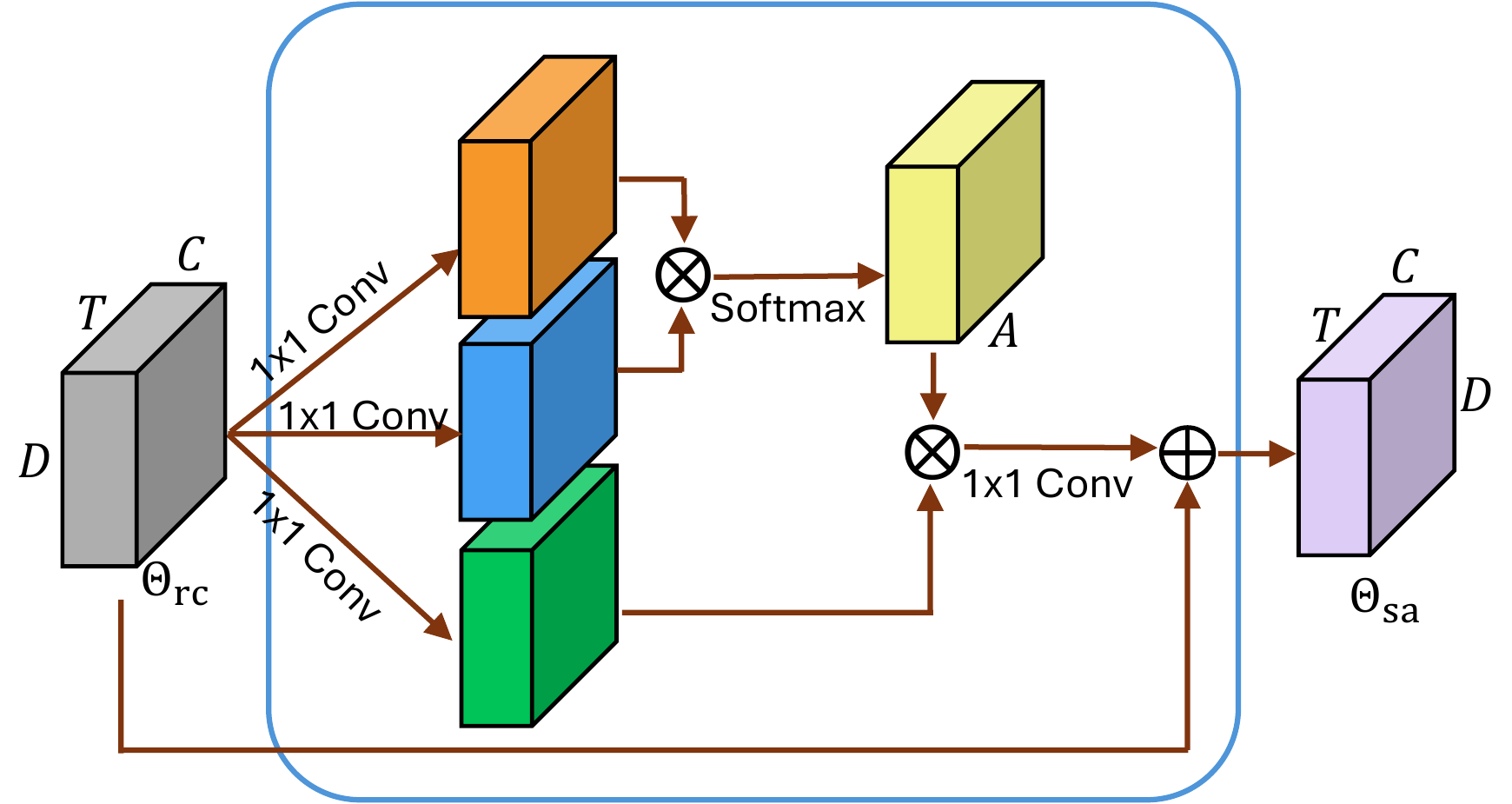}
\caption{Spatial Transformer Block.}
\label{fig:sca_block}
\end{figure}





While channel and spatial attention individually enhance feature representations by focusing on informative channels and salient spatial regions, their integration allows the model to jointly capture where and what to emphasize. We propose a spatial-channel attention (SCA) mechanism that sequentially applies channel and spatial attention to generate a more discriminative feature map.

Specifically, the outputs of the channel attention and spatial attention are fused to form the SCA Block. 

Given the channel attention output $\boldsymbol{s}_{\mathrm{channel}}$and the spatial attention output $\Theta_{\mathrm{sa}}\in \mathbb{R}^{C\times D\times T}$, we combine them through element-wise multiplication, such that each spatial feature is reweighted according to its channel importance:
\begin{equation}
    \Theta_{\mathrm{fused}} = \boldsymbol{s}_{\mathrm{channel}} \odot \Theta_{\mathrm{sa}}
\end{equation}
where $\odot$ denotes element-wise product after broadcasting $\boldsymbol{s}_{\mathrm{channel}}$ across the spatial dimensions. Finally, a residual connection is applied to preserve the original feature representation:
\begin{equation}
    \Theta_{\mathrm{rc}}^{\prime} = \gamma_2\cdot\Theta_{\mathrm{rc}} + \Theta_{\mathrm{fused}}
\end{equation}

To further enhance feature representation, multiple SCA blocks are stacked in sequence. As described above, a single SCA block updates the input feature from $\Theta_{\text{rc}}$ to $\Theta_{\text{rc}}^\prime$. Extending this to $L_1$ stacked blocks, we denote the input to the $l$-th block as $\Theta_{\text{rc}}^{(l)}$, which passed through the SCA block and create an updated feature $\Theta_{\text{rc}}^{(l+1)}$. Let $\Theta_{\text{rc}}^{(0)}$ represent the initial input and $\Theta_{\text{rc}}^{(L_1)}$ be the final refined presentation, then we have
\begin{equation}
    \Theta_{\text{rc}}^{(L_1)} = \text{SCA}^{\circlearrowright L_1} (\Theta_{\text{rc}}^{(0)})
\end{equation}

Finally, a $1\times 1$ convolution is applied to project the feature map into a single channel, yielding the reconstructed CGM profile from the SMBG input:

\begin{equation} 
 \widehat{\boldsymbol{G}}= \text{Conv}_{1\times 1}(\Theta_{\text{rc}}^{(L_1)})
\end{equation}

To ensure accurate trajectory reconstruction and maintain numerical stability, we define a restoration loss that encourages the reconstructed CGM profile $\widehat{\boldsymbol{G}} = \{\hat{g}_{d,t}\mid d=1,\cdots,D; t=1,\cdots,T\}$ to closely match the ground-truth CGM $ \boldsymbol{G} = \{g_{d,t}\mid d=1,\cdots,D; t=1,\cdots,T\}$. This is achieved by minimizing the mean squared error (MSE) between the two trajectories, defined as:
\begin{equation}
    \mathcal{L}_\mathrm{rc} = \frac{1}{DT}\sum_{d=1}^D \sum_{t=1}^T\left\Vert g_{d,t}-\hat{g}_{d,t}\right\Vert^2
\end{equation}

This loss guides the upper path to recover physiologically plausible CGM trajectories from sparse SMBG, which in turn enables the derivation of surrogate AGP metrics

The proposed SCA path learns the pattern from SMBG data and progressively reconstructs a continuous profile by capturing both inter-day and intra-day dependencies uisng the spatial-channel attention mechanism, adaptively emphasizing salient regions and informative channels. This reconstruction not only provides physiologically plausible trajectories but also serves as a foundation for deriving surrogate AGP metrics in a manner consistent with CGM-based computation.

\subsection{Multi-scale ResNet Path for Glycemic Metric Prediction}


The multi-scale ResNet path forms the lower branch of the proposed model and is designed to directly predict the glycemic metrics—$\widehat{\text{TIR}}^{(L)}, \widehat{\mathrm{TAR}}^{(L)}, \widehat{\text{TBR}}^{(L)}$—from sparse SMBG inputs. Unlike the upper SCA path, which reconstructs a continuous glucose trajectory before deriving surrogate AGP metrics, the Multi-scale ResNet path learns a direct mapping from the input representation to the target metrics. 

To achieve direct prediction of glycemic metrics, the lower branch first adopts a ResNet backbone \cite{he2016deep} to efficiently learn and refine high-level feature representations while mitigating vanishing gradients and facilitating stable optimization. Specifically, the ResNet module consists $L_2=4$ sequential residual layers with progressively increasing channel dimensions. Each residual block is composed of two residual layers, and each layer contains two convolutional operations with batch normalization and ReLU activation, together
with an identity skip connection. 

Subsequently, an Atrous Spatial Pyramid Pooling (ASPP) module \cite{chen2017rethinking} is incorporated to further enhance the model’s multi-scale feature extraction capability from SMBG inputs. This design allows the model to directly map sparse glucose observations to clinically meaningful percentage estimates, complementing the reconstruction path by providing outcome-focused supervision. Specifically, the ASPP module consists of multiple parallel branches, which include: 
\begin{itemize}
    \item one $1\times 1$ convolution branch to preserve local features;
    \item three $3\times 3$ atrous convolution braches with different dialation rates $\gamma_d = 6,12,18$ to capture temporal patterns at various scales;
    \item one global average pooling (GAP) branch that captures global context information, followed by a $1 \times 1$ convolution and bilinear interpolation to restore the feature map to the original spatial size.
\end{itemize}

The outputs of all branches are concatenated along
the channel dimension and fused through a $1 \times 1$ convolution, resulting in an intermediate feature map $\Theta_{\text{aspp}}$. Then, we apply GAP to compress $\Theta_{\text{aspp}}$ to a single vector $V$, which is passed through a fully connected layer to produce three final metric prediction logits $\boldsymbol{\eta} \in \mathbb{R}^3$. To ensure that the predicted metrics possess biological plausibility and adhere to the definition of percentage-based measures, the outputs are constrained to satisfy: $ \widehat{\text{TBR}}^{(L)}+ \widehat{\text{TIR}}^{(L)}+ \widehat{\text{TAR}}^{(L)}=1 $, $\{\widehat{\text{TBR}}^{(L)},\widehat{\text{TIR}}^{(L)}, \widehat{\text{TAR}}^{(L)}\}\in [0,1] $. To ensure that the predicted metrics are physiologically interpretable as percentages, the three output logits are normalized using a softmax activation:
\begin{equation}
    \widehat{\text{TR}}^{(L)} = \text{Softmax}(\boldsymbol{\eta})
\end{equation}

As such, the neural network outputs $\widehat{\text{TR}}^{(L)} =\{\widehat{\text{TBR}}^{(L)},\widehat{\text{TIR}}^{(L)}, \widehat{\text{TAR}}^{(L)}\} $ can be directly interpreted as biological meaningful percentage predicted from the lower path. The lower path bypasses trajectory generation and focuses solely on outcome-level prediction. This design enhances efficiency by directly learning the mapping from sparse SMBG inputs to clinically relevant metrics. To supervise these predictions, we define a metric prediction loss $\mathcal{L}_\text{tr}$ that measures the discrepancy between the predicted metrics $\widehat{\text{TR}}^{(L)} $ and the ground-truth glycemic AGP metrics $\text{TR} = \{\text{TBR},  \text{TIR} , \text{TAR}\}$. We employ the MSE loss:

\begin{equation}
    \mathcal{L}_\text{tr} = \frac{1}{3}\sum_{j\in \{\text{TBR},  \text{TIR} , \text{TAR}\}}\| \text{TR}_j - \widehat{\text{TR}}_j^{(L)} \| ^2
\end{equation}

This loss encourages the Multi-scale ResNet path to learn a direct mapping from sparse SMBG inputs to clinically meaningful glycemic metrics.




\begin{figure*}[htbp]
\centering
\includegraphics[width=0.75\textwidth]{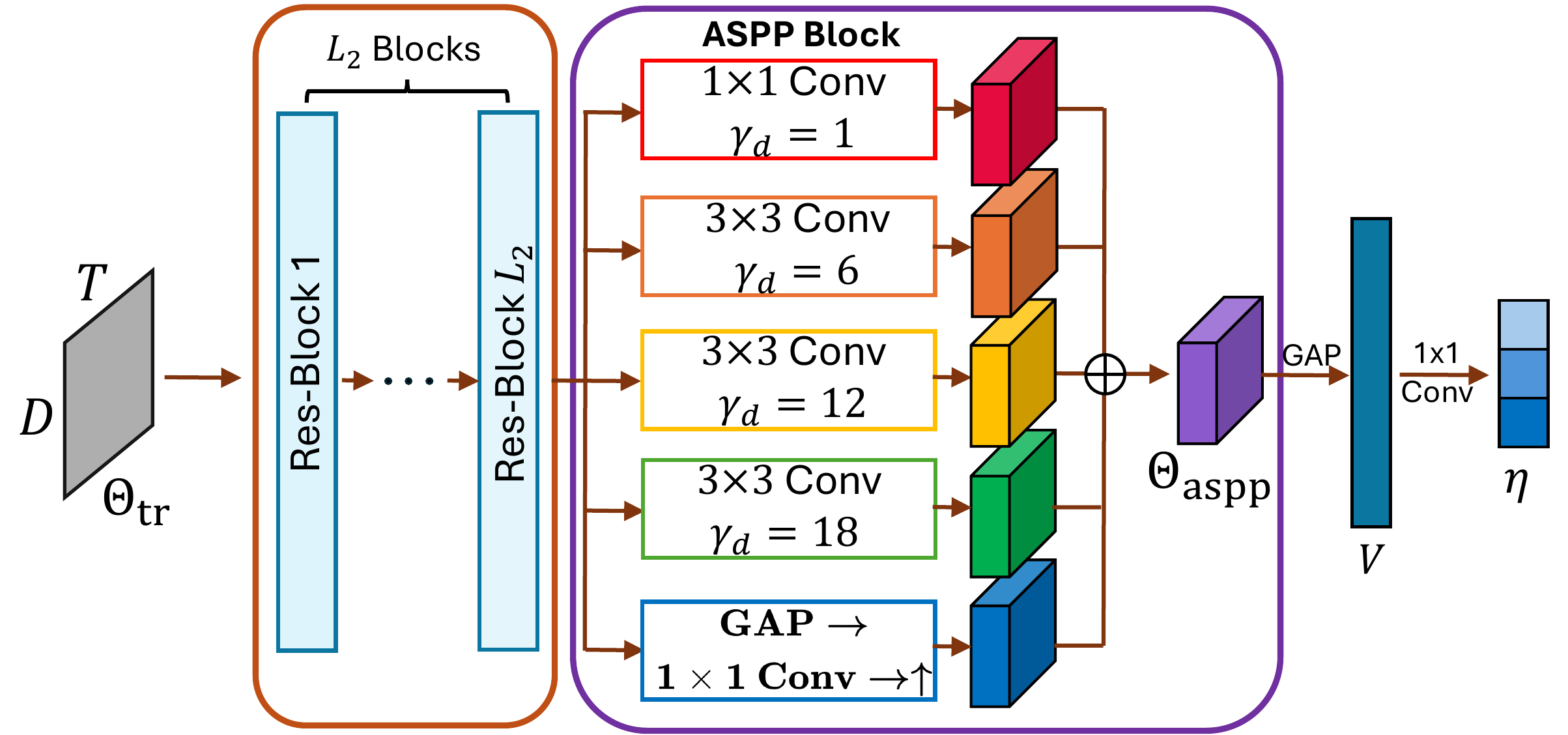}
\caption{Multi-scale ResNet Path.}
\label{fig:resnet_aspp}
\end{figure*}

\subsection{Path Alignment for Consistent Prediction}








Since the dual-path architecture captures complementary aspects of SMBG patterns—the reconstruction path modeling continuous glucose trajectories and the direct path focusing on outcome-level prediction—it is desirable for both branches to produce consistent estimates of the AGP metrics. In particular, although the two paths differ in their learning objectives, the values of TBR, TIR, and TAR derived from the reconstructed CGM should remain close to those predicted directly by the Multi-scale ResNet branch. To enforce this consistency, we introduce an alignment loss that constrains the AGP metrics from the upper path to be aligned with those from the lower path. 

Specifically, the derivation of AGP metrics from the upper path follows the same procedure as the ground-truth computation, by calculating the fraction of reconstructed CGM values that fall within each glycemic region, as described in Eq. \ref{Eq:cgm_to_tr}. Once the surrogate metrics are obtained from the reconstruction, i.e., $\widehat{\text{TR}}^{(U)} =\{\widehat{\text{TBR}}^{(U)},\widehat{\text{TIR}}^{(U)}, \widehat{\text{TAR}}^{(U)}\} $, we enforce the consistency with the direct predictions from the lower path through an alignment loss defined as the mean squared error (MSE) between the two:

\begin{equation}
\mathcal{L}_{\mathrm{a}} = \frac{1}{3} \sum_{j \in {\mathrm{TBR}, \mathrm{TIR}, \mathrm{TAR}}} \| \widehat{\text{TR}}^{(U)}_j - \widehat{\text{TR}}^{(L)}_j \|^2
\end{equation}

This loss constrains the AGP metrics derived from the reconstructed CGM to remain close to those predicted directly, thereby ensuring that both paths contribute consistent and complementary information. The total loss function is then formulated as a weighted combination of the reconstruction loss, the direct prediction loss, and the alignment loss:
\begin{equation}
\mathcal{L}_{\text{total}}
= \lambda_{\mathrm{rc}}  \mathcal{L}_{\mathrm{rc}}
+ \lambda_{\mathrm{tr}} \mathcal{L}_{\mathrm{low}}
+ \lambda_{\mathrm{a}} , \mathcal{L}_{\mathrm{a}}
\end{equation}

where $\lambda_\mathrm{rc}, \lambda_\mathrm{tr}, \lambda_\mathrm{a} \geq 0$ are hyperparameters that balance the contributions of the three terms. This composite objective encourages the model to simultaneously reconstruct physiologically plausible CGM trajectories, accurately predict glycemic metrics directly from SMBG, and maintain consistency between the two paths.

\subsection{Active Point Selector}
\label{section:active_selector}
In real-world settings, the timing of blood glucose measurements is often guided by patient behavior and physiological conditions rather than random processes\cite{parkin2009value,choudhary2013blood,patton2015adherence}. For instance, patients are more likely to perform fingerstick tests when experiencing symptoms of hypoglycemia, or around mealtimes (before and after meals). Conversely, measurements are much less likely to occur during sleep hours, such as from midnight to 5 a.m., when patients are typically not awake or actively monitoring their glucose levels.

This behavioral pattern suggests that SMBG data exhibit strong temporal biases\cite{namayanja2012assessment}, and any model that aims to simulate SMBG observations should account for these preferences. In contrast, using random sampling to mimic SMBG measurements may overlook clinically relevant time points, potentially reducing model performance and interpretability\cite{polonsky2011structured}.
For example, measurements are most frequent during morning and post-meal hours, while drastically fewer samples are collected during the night. This motivates the development of an active point selector that can learn to identify the most informative and behaviorally plausible time points for downstream modeling.


To identify representative observation points from CGM sequences, we developed an active point selection model based on a multi-scale, attention-enhanced temporal convolutional architecture, hereafter referred to as the AETCN selector. This selector operates independently from the dual-path attention network training. It relies solely on the paired data collected from patients who simultaneously underwent CGM monitoring and SMBG testing. Based on this supervised training, the AETCN Selector assigns a probability score to each time step and selects a subset of time points that are most likely to serve as informative SMBG measurements from the full CGM sequence. The selected SMBG points, together with their corresponding CGM values, can then be used to formulate a trustworthy training set for models that estimate AGP metrics from sparse SMBG observations. Please refer to Appendix \ref{sec:appendix} for the detailed network structure.


\section{EXPERIMENTAL DESIGN AND RESULTS}

\subsubsection{Data Source}
In this study, we utilized publicly available datasets provided by the Jaeb Center for Health Research through the Jaeb Center Public Data Repository for diabetes research. The repository includes CGM and SMBG data collected from individuals with diabetes across multiple clinical studies.
Among these datasets, a specific subset referred to as the \texttt{repbg} cohort contains both CGM and self-monitoring of blood glucose (SMBG) data. This dataset serves two purposes in our study: (i) it provides labeled SMBG events that allow us to analyze the temporal patterns of SMBG actions (i.e. when patients perform fingerstick measurements), and (ii) it supports validation of the proposed DPA-Net model. In addition to the \texttt{repbg} cohort, other CGM-based datasets from the repository were used for training DPA-Net. In this setting, the model input consists of SMBG samples that are either randomly selected or identified by the active point selector model (see Section \ref{section:active_selector}) from CGM, meanwhile the ground-truth AGP metrics can be directly computed from the corresponding CGM sequences. The overall task is formulated as supervised learning, where the goal is to estimate two-week AGP metrics ($\widehat{\text{TR}}$) from sparse SMBG observations.

\subsubsection{Label Definition}

In this study, we define the ground-truth labels for training and evaluation based on three high-level glycemic ranges:  Time In Range (TIR), Time Above Range (TAR), and Time Below Range (TBR), which are calculated from Eq. \ref{Eq:cgm_to_tr}. The SMBG AGP Report  uses 14 days of SMBG data to provide point estimates of time in ranges metrics, which are reported using the “thermometer” as shown in Fig. \ref{fig:tr_description}.  
For each sample, we use the $D=14$ days CGM record $\boldsymbol{G} \in \mathbb{R}^{D \times T}$ to calculate the ground-truth AGP metrics $\text{TR}$, which will be served as the target labels against which the model predictions $\widehat{\text{TR}}$ are evaluated.

\subsection{Evaluation Metrics}
The performance of the models is evaluated separately for each prediction target using the root mean squared error (RMSE), the coefficient of determination ($R^2$), and the mean absolute error (MAE). The primary evaluation is conducted on the \texttt{repbg} dataset, which contains real SMBG measurements from a subset of patients, thereby providing a realistic benchmark for assessing model generalization under actual sampling patterns.

Let $\text{TR}^{(j)}_\rho$ denote the ground-truth value of the $j$-th target metric ($j \in \{\mathrm{TAR}, \mathrm{TIR}, \mathrm{TBR}\}$) for the $\rho$-th patient, and $\hat{\text{TR}}^{(j)}_\rho$ denote the corresponding prediction, where $N$ is the total number of samples in the evaluation set. The metrics are defined as:

\begin{align}
\mathrm{RMSE}^{(j)} &= \sqrt{\frac{1}{N} \sum_{\rho=1}^{N} \left(\text{TR}^{(j)}_\rho - \widehat{\text{TR}}^{(j)}_{\rho}\right)^2 \nonumber}\\
\mathrm{R}^{2(j)} &= 1 - \frac{\sum_{\rho=1}^{N} \left(\text{TR}^{(j)}_\rho - \widehat{\text{TR}}^{(j)}_\rho\right)^2}{\sum_{\rho=1}^{N} \left(\text{TR}^{(j)}_\rho - \overline{\text{TR}}^{(j)}\right)^2}\nonumber\\
\mathrm{MAE}^{(j)}  &= \frac{1}{N} \sum_{\rho=1}^{N} \left|\text{TR}^{(j)}_\rho - \widehat{\text{TR}}^{(j)}_{\rho}\right| \nonumber\\
\end{align}

where $\overline{\text{TR}}^{(j)}$ denotes the mean of the ground-truth values for the $j$-th target. 
The overall $RMSE$ and $R^2$ are reported as the average across the three targets, i.e., 
\begin{align}
\mathrm{RMSE} &= \frac{1}{3}\sum_{j\in \mathrm{TAR, TIR, TBR}} \mathrm{RMSE}^{(j)} \nonumber\\
\mathrm{R}^2 &= \frac{1}{3}\sum_{j\in \mathrm{TAR, TIR, TBR}} \mathrm{R}^{2(j)}\nonumber\\
\end{align}

Lower values of RMSE and MAE (closer to 0) indicate better predictive performance, while higher values of $R^2$ (closer to $1$) indicate stronger goodness-of-fit.





\subsection{Active Point Selection}
\label{sec:active_point_selection}


First, we evaluate the reliability of the potential SMBG points suggested by the active point selector model, which are derived from CGM data. In real-world settings, patients often follow certain behavioral or daily routines that trigger finger-stick SMBG measurements. For example, patients may check their glucose before breakfast, around mealtimes, at bedtime, or when they feel symptoms such as dizziness or palpitations. The active selector aims to capture such patterns adaptively, thereby tailoring the prediction of likely SMBG events. Since the dataset consists primarily of CGM data without corresponding SMBG records, the abiThe objective of the active selector is to identify the most likely SMBG points by assigning a score to each time step from CGM patterns and selecting those with the highest scores. The active selector model is trained on \texttt{repbg} dataset, where ground truth CGM and SMBG are both included, laying the data foundation for training active point selector in a supervised manner. During inference, we fix the number of selected SMBG points to 5 per day for simplicity and comparability.

\begin{figure}
\centering
\includegraphics[width=1.0\linewidth]{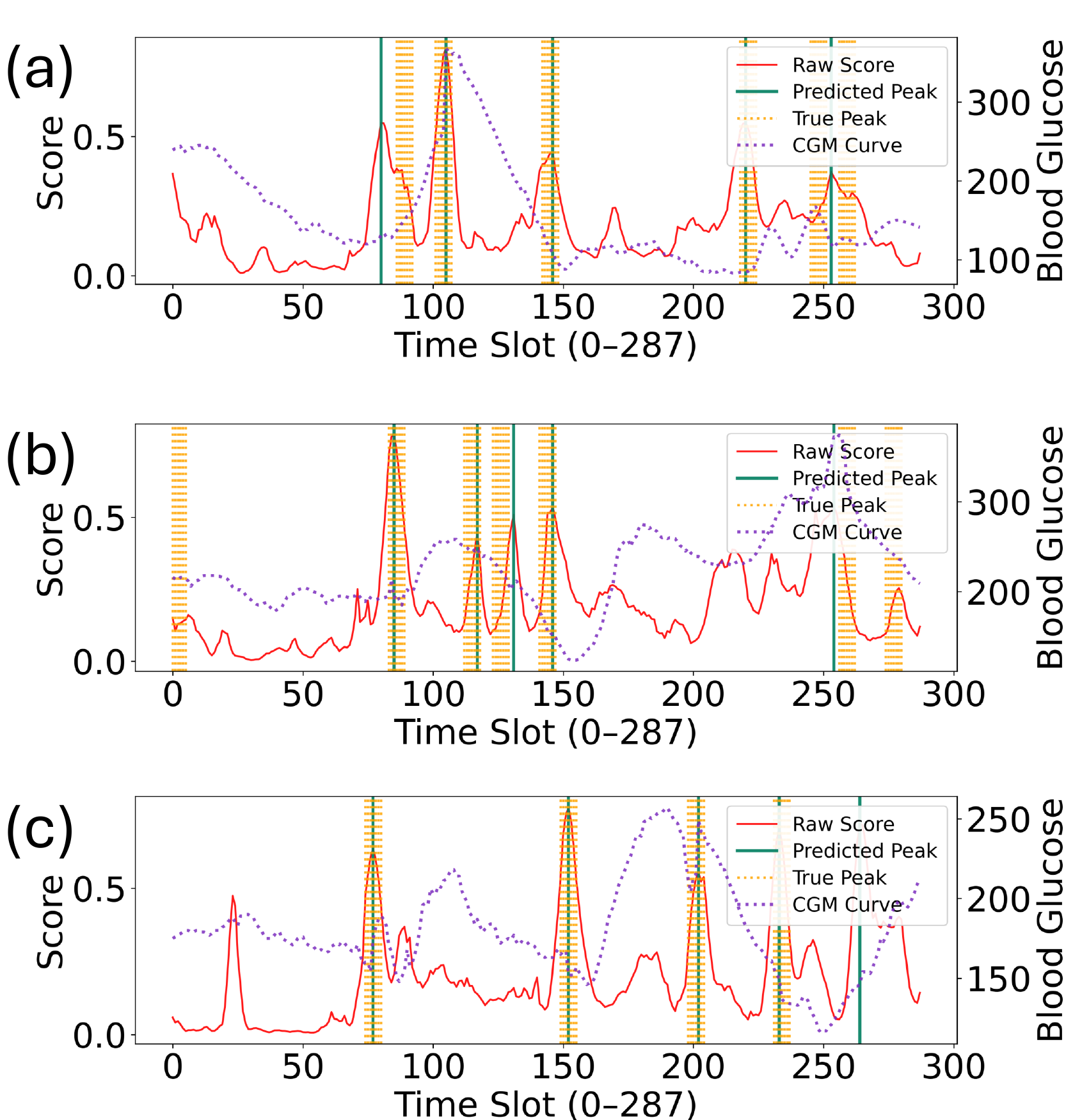}
\caption{Three examples of point selection by the Active Selector.}
\label{fig:active_selector_examples}
\end{figure}

Fig. \ref{fig:active_selector_examples} illustrates the point selection results of the active point selector. In each example, the purple dotted curve represents the ground-truth CGM trajectory. The red curve represents the probability distribution over time points, indicating how likely each location is to be selected as an SMBG measurement by the active point selector. The green vertical lines indicate the predicted peak positions, and the yellow shaded regions around the dashed lines denote the vicinity of true glucose peaks (peak $\pm$ tolerance window).

Across different patients and days, the predicted peaks generally align well with the true peaks, demonstrating the robustness and effectiveness of the active point selector. For example, in Fig.~\ref{fig:active_selector_examples}(a), the number of actual SMBG instances on that day is approximately 5. The predicted peaks align closely with the true peaks, accurately capturing the major glucose variations. In Fig.~\ref{fig:active_selector_examples}(b), when the actual number of SMBG observations exceeds $n=5$ in a day, the active point selector still identifies the major peaks, leaving a few SMBG temporal locations with low probabilities not captured due to the limit of SMBG selection number. In Fig.~\ref{fig:active_selector_examples}(c), as the number of true peaks is fewer than the selected number, predicted peaks almost perfectly coincide with the true peaks. The active point selector additionally chooses some time points it considers worth measuring.



It is also worth noting that the true SMBG measurement times are highly subjective in practice. Clinically, individuals tend to measure glucose more frequently when levels are dropping, as hypoglycemia is often accompanied by uncomfortable symptoms. This trend can also be observed in our examples, where true peaks are more likely to fall within glucose-decreasing regions. This observation further explains why the active point selector sometimes proposes additional time points beyond the true SMBG measurements: these extra selections often fall near rapid declines or potential hypoglycemic episodes, making them clinically reasonable and informative.

In summary, although the number of SMBG measurements per day is random and varies across patients, the active selector can consistently capture clinically meaningful glucose fluctuations and, when necessary, propose additional candidate time points for measurement that are both reasonable and potentially valuable in practice.



\subsection{Impact of SMBG Point Selection on Model Performance}

In Section~\ref{sec:active_point_selection}, we demonstrated the effectiveness of the active point selector in identifying potential SMBG instances among CGM time series.In this section, we analyze how different SMBG selection strategies for constructing the training set influence the prediction performance of DPA-Net. 

In particular, two major types of strategies are introduced. The first strategy is random masking, in which SMBG points are sampled randomly from CGM data at a predefined selection rate. The second strategy is a hybrid selection scheme, parameterized by the selector ratio $\gamma_h$, in which a $\gamma_h$ fraction of days that use the active point selector to choose SMBG points (e.g., $n=5$ per day), while the remaining $(1-\gamma_h)$ fraction of days use random selection. The hybrid strategy combines the adaptiveness of active point selector, which reflects real behavioral patterns, with the diversity of random sampling to avoid bias from fixed behavioral routines.
Using the selected SMBG points as training inputs to the proposed DPA-Net, we report its comparative performance on the unseen test set in Table~\ref{tab:masking_comparison}.



\begin{table*}[htbp]
\centering
\scriptsize
\setlength{\tabcolsep}{8pt}
\renewcommand{\arraystretch}{1.2}
\begin{tabular}{|l|c|c|c|c|c|c|c|}
\hline
\text{Selection Strategy} & \text{$\gamma_h$} & \text{Selection Rate} & \text{Overall RMSE} & \text{Overall $R^2$} & \text{TAR MAE} & \text{TIR MAE} & \text{TBR MAE} \\
\hline
Random              & --   & 30\%   & 0.1881 & -0.9708 & 0.1907 & 0.2077 & 0.0232 \\
Random              & --   & 20\%   & 0.0755 & 0.5705  & 0.0702 & 0.0669 & 0.0170 \\
Random              & --   & 10\%   & 0.0762 & 0.3631  & 0.0760 & 0.0660 & 0.0235 \\
Random              & --   & 5\%   & 0.0624 & 0.5090  & 0.0553 & 0.0600 & 0.0209 \\
Random              & --   & 2.8\%(8) & 0.0608 & 0.6543  & 0.0576 & 0.0573 & 0.0153 \\
Random              & --   & 2.5\%(7) & 0.0806 & 0.4683  & 0.0759 & 0.0779 & 0.0183 \\
Random              & --   & 2.1\%(6) & 0.0824 & 0.4910  & 0.0819 & 0.0749 & 0.0165 \\
Hybrid              & 0.2  & 2.8\% & 0.0693 & 0.2758  &   0.0680 & 0.0594 & 0.0273 \\
Hybrid              & 0.3  & 2.8\% & 0.0721 & 0.5776  & 0.0708 & 0.0687 & 0.0160 \\
Hybrid              & 0.35 & 2.8\% & 0.0721 & 0.5533  & 0.0739 & 0.0655 & 0.0172 \\
Hybrid              & 0.4  & 2.8\% & 0.0500 & 0.6912  & 0.0476 & 0.0465 & 0.0163 \\
Hybrid              & 0.45 & 2.8\% & 0.0648 & 0.4878  & 0.0659 & 0.0571 & 0.0215 \\
Hybrid              & 0.5  & 2.8\% & 0.0700 & 0.5832  & 0.0704 & 0.0658 & 0.0165 \\
Hybrid              & 0.6  & 2.8\% & 0.0640 & 0.5704  & 0.0639 & 0.0579 & 0.0174 \\
Hybrid              & 0.7  & 2.8\% & 0.0632 & 0.6019  & 0.0601 & 0.0571 & 0.0173 \\
Hybrid              & 0.8  & 2.8\% & 0.0721 & 0.5725  & 0.0715 & 0.0663 & 0.0165 \\
\hline
\end{tabular}
\caption{Performance of the dual-path model under different masking strategies on the \textit{repbg} test set. Reported metrics include overall RMSE and $R^2$, and MAE for TAR/TIR/TBR.}
\label{tab:masking_comparison}
\end{table*}

From Table~\ref{tab:masking_comparison}, it can be observed that the performance of the model under random selection does not change monotonically with the rate of random selection. Even trained with high SMBG selection rates (30\%), the overall performance is notably poor. 
This is caused by a distribution mismatch between training and testing:The SMBG counts in the \texttt{repbg} test set are highly sparse, with selection rates of only about 1–4\% relative to the full CGM collections. When training is conducted with SMBG selection rate of 30\%, the sparsity level seen during training deviates substantially from that of the test set, leading to poor generalization on real \texttt{repbg} dataset. In contrast, when the SMBG training selection rate is reduced to 5-20\%, the sparsity level of the training set becomes closer to that of the test set, resulting in a remarkable improvement in evaluation performance. Further improvements are achieved at 3-5\% SMBG selection, where the model reaches its best overall performance. However, when the SMBG selection rate for training approaches or even falls below the test set’s mean selection rate (2.5\%), performance drops again, indicating that extreme sparsity introduces instability.

In random selection with varying SMBG sampling rates as the training input, the model achieves its best performance at 2.8\% ($\approx$8 points per day). We therefore take this setting as the baseline and design the hybrid selection strategy upon it: for each sample, random selection with a fixed 2.8\% rate is applied to $(1-\gamma_h)$ of the days, while the active selector (fixed at 5 points per day) is applied to the remaining $\gamma_h$ days. Results show that the overall performance of the hybrid strategy fluctuates as $\gamma_h$ varies, with the best performance achieved at $\gamma_h = 0.4$, yielding the lowest RMSE, MAE, and highest $\text{R}^2$. This indicates that introducing a moderate degree of active selection can significantly enhance generalization. In contrast, an excessively high ratio of $\gamma_h$ may cause the model to learn overly rigid patterns and reduce data diversity, while a too low ratio prevents the contribution of the active selector from being effectively leveraged.

\begin{figure}[htbp]
    \centering

    \begin{subfigure}[t]{1.0\linewidth}
        \centering
        \includegraphics[width=\linewidth]{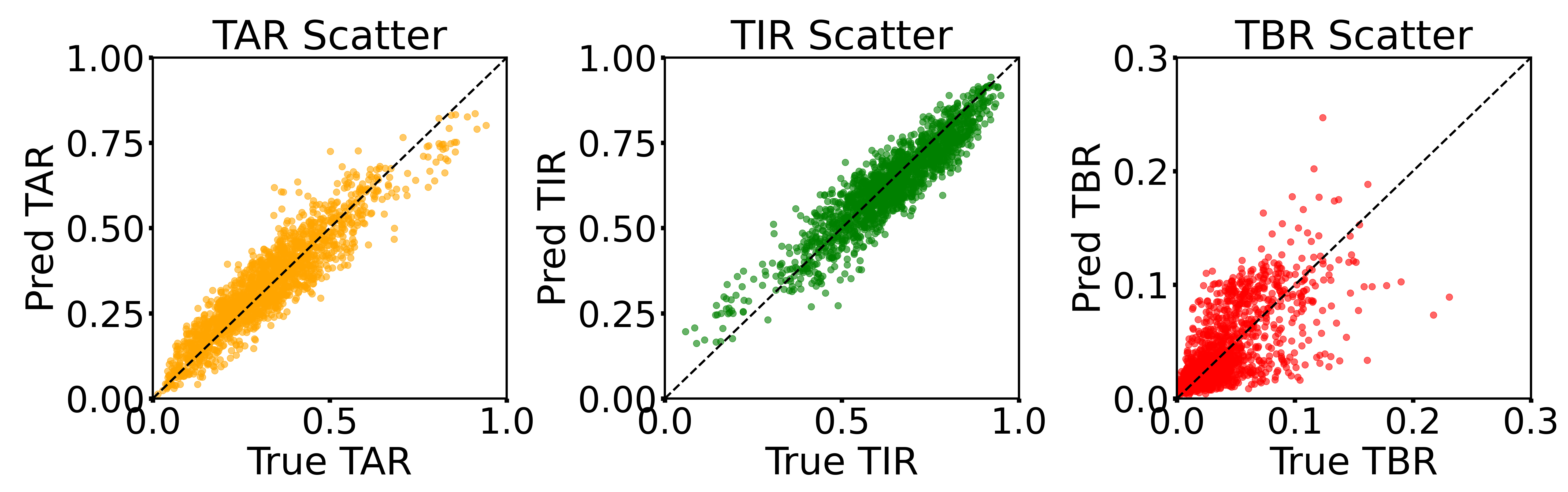}
        \caption{Hybrid ($\gamma_h = 0.4$; best)}
    \end{subfigure}
    \vspace{0.2cm}

    \begin{subfigure}[t]{1.0 \linewidth}
        \centering
        \includegraphics[width=\linewidth]{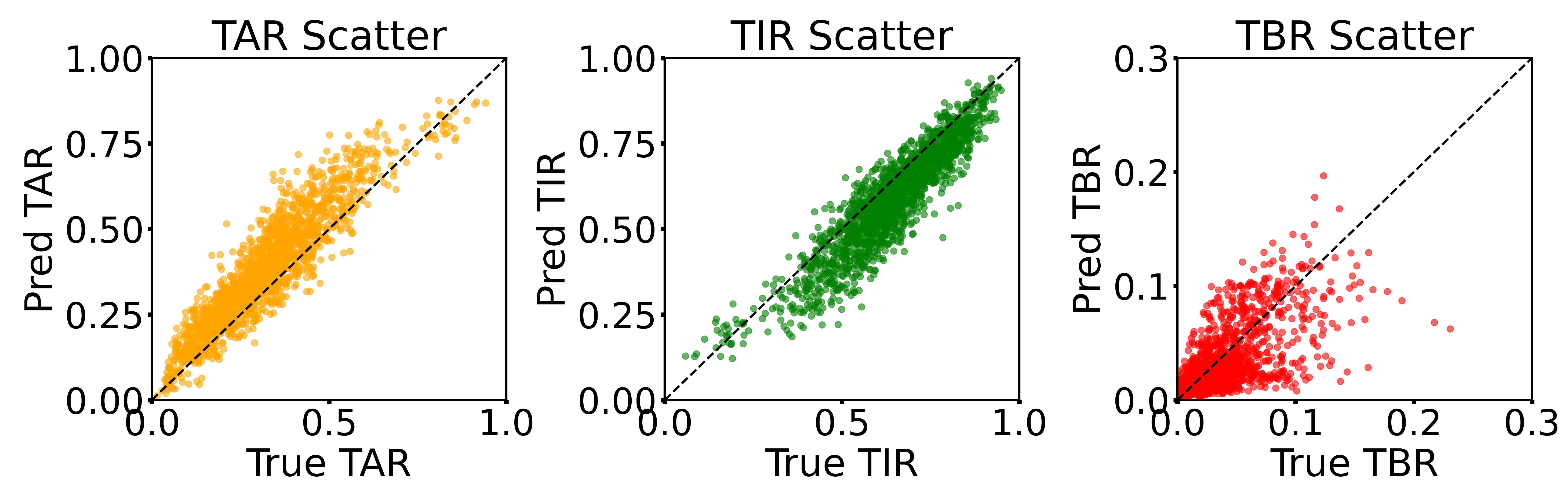}
        \caption{Hybrid ($\gamma_h = 0.8$)}
    \end{subfigure}

    \begin{subfigure}[t]{1.0\linewidth}
        \centering
        \includegraphics[width=\linewidth]{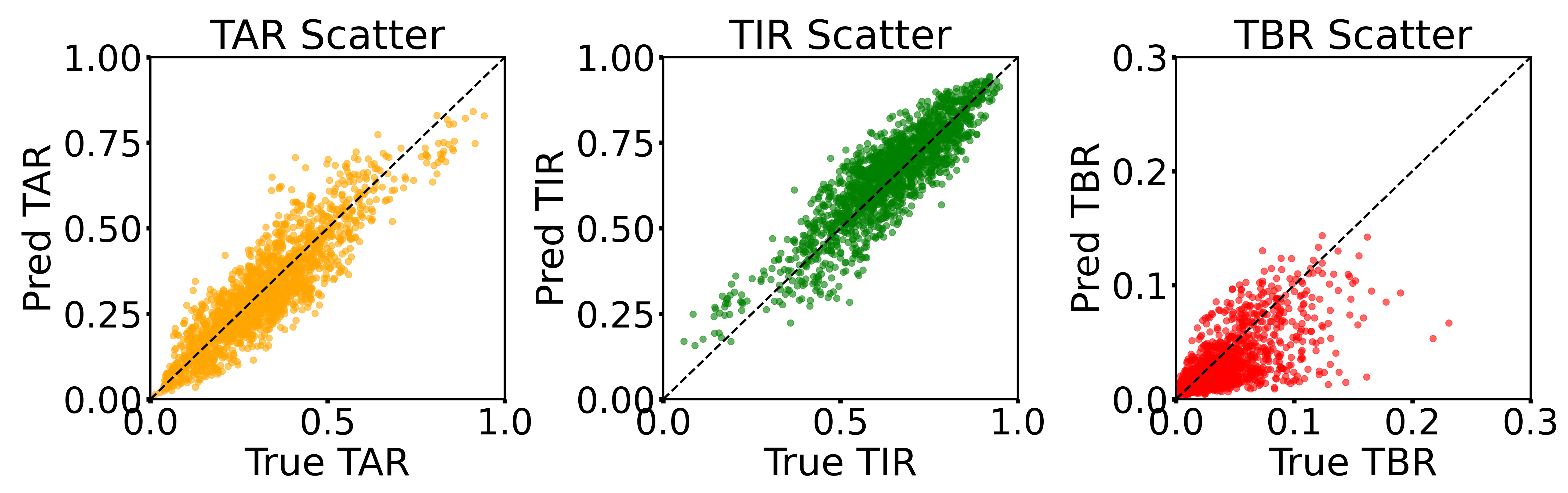}
        \caption{Random Selection (2.8\%)}
    \end{subfigure}
    \vspace{0.2cm}
    
    \caption{Predicted vs. True scatterplots on the test set under three masking strategies (top to bottom): 
    and hybrid($\gamma_h=0.4$), hybrid($\gamma_h=0.8$), and random (2.8\%)). 
    Each panel overlays the identity line ($y{=}x$).}
    \label{fig:selection_strategy_scatter}
\end{figure}

To visually substantiate these findings, 
Fig.~\ref{fig:selection_strategy_scatter} presents scatter plots of predicted versus true AGP metrics (TAR, TIR, and TBR) under different SMBG selection strategies. Fig.~\ref{fig:selection_strategy_scatter}(a) shows the hybrid strategy with $\gamma_h = 0.4$, which yields the best performance; Fig.~\ref{fig:selection_strategy_scatter}(b) illustrates the hybrid strategy with $\gamma_h = 0.8$; and Fig. \ref{fig:selection_strategy_scatter}(c) depicts the baseline random selection strategy (2.8\%).
From Fig. \ref{fig:selection_strategy_scatter}, the hybrid selection strategy ($\gamma_h= 0.4$) produces TAR and TIR point clouds that adhere more closely to the diagonal, indicating strong agreement between predicted and true values. 
 While TBR remains more scattered across all strategies due to its relatively low prevalence, the $\gamma_h = 0.4$ setting still provides the most de-biased estimates among the compared approaches.

Compared with the hybrid selection strategy at $\gamma_h=0.8$, the setting with $\gamma_h=0.4$ markedly reduces systematic bias across TAR, TIR, and TBR. Specifically, $\gamma_h=0.8$ leads to overestimation of TAR and underestimation of TIR and TBR, whereas $\gamma_h=0.4$ produces predictions that are more closely aligned with the diagonal. Similarly, under the pure random scheme with a selection rate of $2.8\%$, the biases in TIR and TAR are less pronounced, but the point clouds display greater dispersion from the diagonal compared with the hybrid strategy with $\gamma_h=0.4$. The TBR scatter plot reveals a tendency to underestimate relative to the ground truth.

Overall, hybrid selection with selector ratio $\gamma_h=0.4$ provides the best-calibrated and least-dispersed predictions across TAR/TIR/TBR among the three selection strategies.



\subsection{Model Validation and Ablation of DPA-Net}

In the current investigation, we systematically evaluate the efficacy of the DPA-Net framework through ablation studies and comparative analyses, validating its predictive performance in estimating AGP glycemic metrics from 2-week SMBG data.
As illustrated in Fig.~\ref{fig:overall_dualpath}, we evaluate two reduced model variants and compare them against the full DPA-Net:
(1) \textbf{Lower path only}: only the multi-scale ResNet path is retained, directly predicting the three AGP glycemic metrics;
(2) \textbf{Upper path only}, only the spatial–channel attention path is retained, reconstructing a CGM-like trajectory from SMBG observations;
The detailed results are reported in Table~\ref{tab:ablation}, showing that both paths contribute complementary benefits, bridged by the alignment loss.

\begin{table*}[htbp]
\centering
\small
\setlength{\tabcolsep}{6pt}
\renewcommand{\arraystretch}{1.2}
\begin{tabular}{|l|c|c|c|c|c|}
\hline
\text{Model Variant} & \text{CGM RMSE} & \text{$\text{R}^2$} & \text{TAR MAE} & \text{TIR MAE} & \text{TBR MAE} \\
\hline
Multi-scale ResNet Path Only              & 0.1112 & 0.0528 & 0.1262 & 0.1063 & 0.0241 \\
SCA Path Only         & 0.3562 & -5.3608 & 0.4043 & 0.3913 & 0.0358 \\
Full Dual Path (Ours)         & 0.0500 & 0.6912  & 0.0476 & 0.0465 & 0.0163 \\
\hline
\end{tabular}
\caption{Ablation study of the dual-path model on the \textit{repbg} test set. Results demonstrate the contribution of each path and the alignment loss to TR prediction.}
\label{tab:ablation}
\end{table*}

\begin{figure}[htbp]
    \centering

    \begin{subfigure}[t]{1.0\linewidth}
        \centering
        \includegraphics[width=\linewidth]{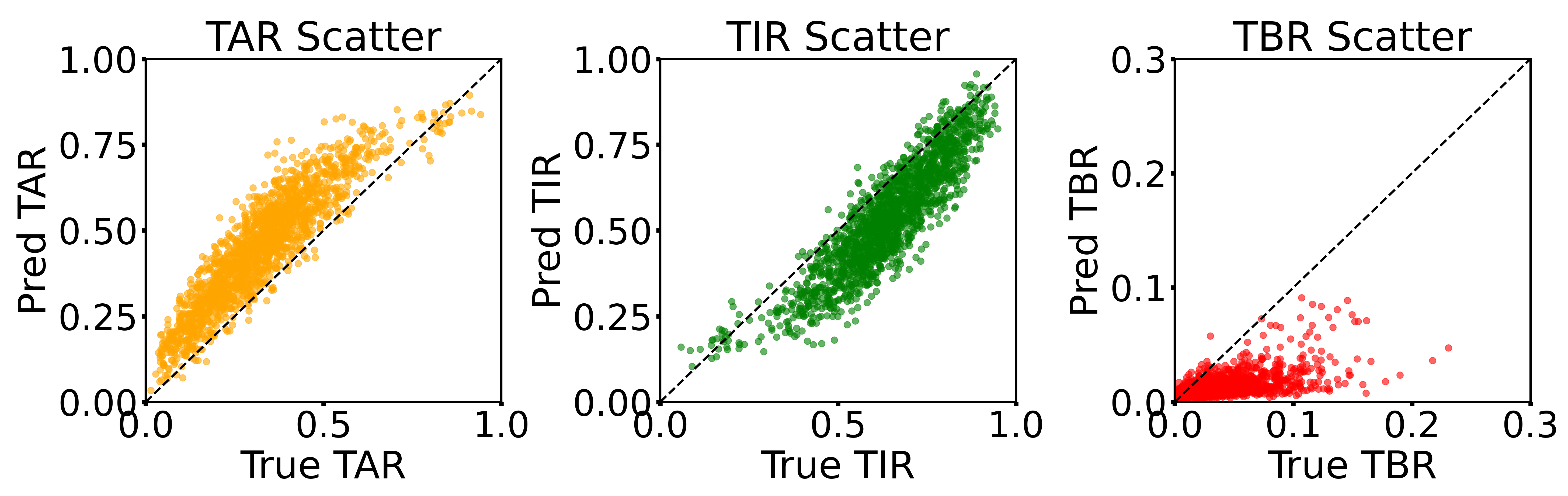}
        \caption{Multi-scale ResNet only}
    \end{subfigure}
    \vspace{0.2cm}

    \begin{subfigure}[t]{1.0 \linewidth}
        \centering
        \includegraphics[width=\linewidth]{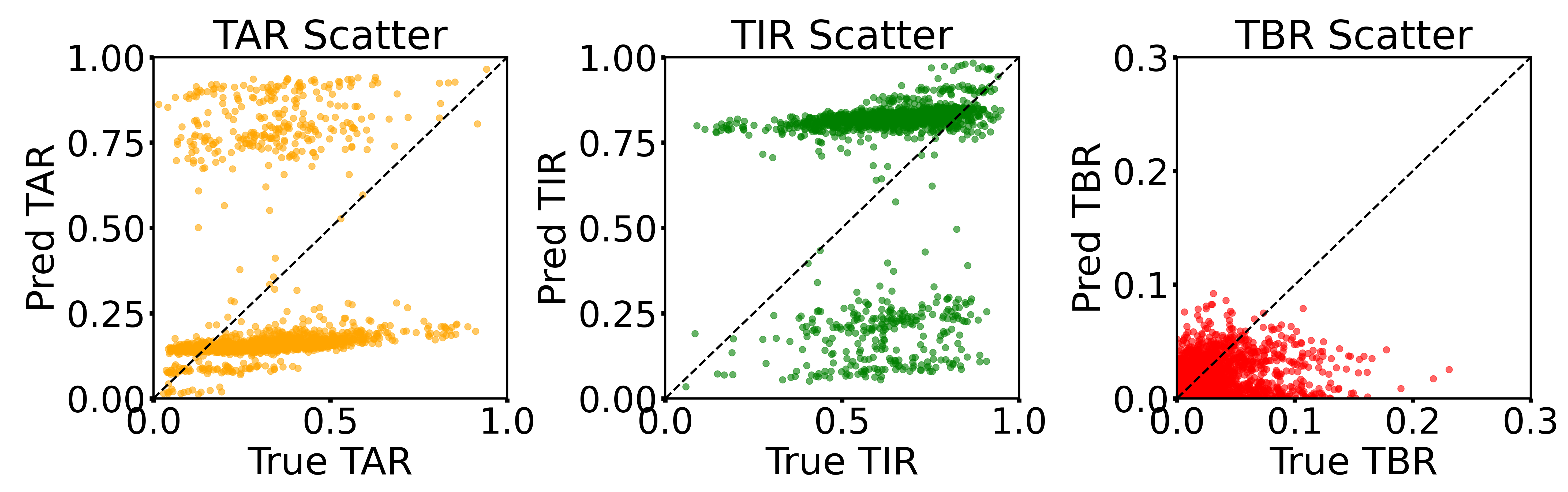}
        \caption{SCA-only}
    \end{subfigure}


    \begin{subfigure}[t]{1.0\linewidth}
        \centering
        \includegraphics[width=\linewidth]{figures/masking_strategy/hybrid0.4.png}
        \caption{Full dual-path (ours)}
    \end{subfigure}
    \vspace{0.2cm}
    
    \caption{Predicted vs. true scatter plots of AGP glycemic metrics for different model variants in the ablation study. 
    Each panel overlays the identity line ($y{=}x$). 
    The full dual-path model shows the tightest clustering around the identity line, while removing one path or the alignment loss introduces bias and dispersion.}
    \label{fig:ablation_scatter}
\end{figure}
From the quantitative results in Table~\ref{tab:ablation}, it can be observed that the superior performance arises from the synergistic effect of the two paths and their strategic alignment in predicting AGP metrics. To intuitively illustrate these results, Fig.~\ref{fig:ablation_scatter} presents scatter plots of the model variants, comparing the distribution of predicted values against the ground truth. From the scatter plots in Fig. \ref{fig:ablation_scatter}, we can further observe the qualitative differences among the ablated variants:

\begin{itemize}
    \item \textbf{Multi-scale ResNet only}: Exhibits substantial systematic bias, with TAR consistently overestimated and TIR/TBR underestimated. Moreover, the scatter distribution is more dispersed compared to the full dual-path model, indicating reduced stability and accuracy.  

    \item \textbf{SCA-only}: This variant fails to provide accurate predictions for TAR and TIR, with scatter points deviating markedly from the identity line, indicating that relying solely on the attention path cannot yield reliable potential CGM trajectory or accurate estimates of AGP metrics. Two potential factors may explain this limitation. First, SMBG is extremely sparse, typically with only about five measurements per day, whereas reconstructing a CGM-like trajectory requires 288 points at 5-minute intervals—representing less than 2\% blood glucose observation coverage. Second, the dataset contains a limited number of patient data, further constraining the model’s ability to generalize.

    \item \textbf{DPA-Net}: 
    In DPA-Net, the alignment between the two paths plays a central role. The spatial–channel attention (SCA) path reconstructs a CGM-like trajectory from SMBG, while being constrained by the AGP metric predictions to ensure consistency in the proportion of time spent within each glycemic range. Conversely, the multi-scale ResNet path, which directly predicts AGP metrics, is adjusted through feedback from the SCA path to better match the reconstructed trajectory. This bidirectional alignment fosters a synergistic effect that reduces potential bias and leads to improved predictive performance.
\end{itemize}

In summary, the full dual-path model achieves the best performance among all variants, demonstrating that both paths and the alignment design are indispensable components of the architecture. 

Since, to the best of our knowledge, no existing approaches estimate AGP metrics from SMBG data, there are no benchmark strategies against which to directly compare our model. We therefore focus on internal comparisons across different SMBG selection schemes and ablation studies to validate the efficacy of our approach.

\subsection{Baselines and Overall Performance}

We establish a simple SMBG baseline (No-Interp) by directly computing the proportions of self-monitored glucose points that fall within the clinical thresholds for TAR, TIR, and TBR, without any temporal interpolation between measurements. The corresponding ground-truth values of {TAR, TIR, TBR} are derived from the CGM data.



Table \ref{tab:baseline_overall} summarizes the overall performance comparison. The No-Interp baseline exhibits noticeable bias due to sparse sampling, particularly for TBR, whereas the proposed DPA-Net markedly reduces errors across all three metrics.

\begin{figure}[t]
\centering
\begin{subfigure}[t]{\linewidth}
  \centering
  \includegraphics[width=\linewidth]{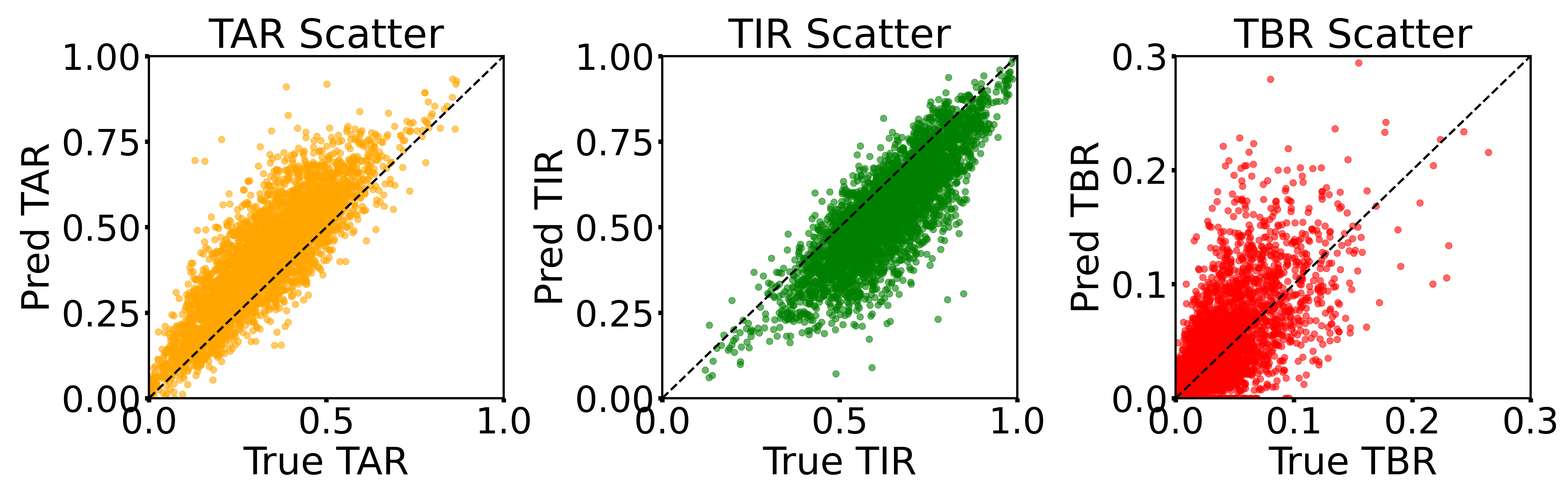}
  \caption{SMBG baseline.}
  \label{fig:scatter-baseline-vert}
\end{subfigure}

\vspace{0.2cm}

\begin{subfigure}[t]{\linewidth}
  \centering
  \includegraphics[width=\linewidth]{figures/masking_strategy/hybrid0.4.png} 
  \caption{Full dual-path (ours).}
  \label{fig:baseline-scatter}
\end{subfigure}
\caption{Predicted vs. true AGP glycemic metric scatter plots for baseline and our method.}
\label{fig:scatter-compare-vert}
\end{figure}

Figure \ref{fig:baseline-scatter} presents the predicted-versus-true scatter plots. As shown, our model consistently outperforms the SMBG No-Interp baseline, achieving lower MAE and RMSE for TAR, TIR, and TBR, thereby demonstrating substantial gains in predictive accuracy. In the scatter plots, DPA-Net’s predictions cluster closely around the diagonal, indicating reduced dispersion and diminished systematic error. In contrast, the baseline tends to overestimate TAR and underestimate TIR, while our model effectively corrects these biases and delivers more accurate, stable, and reliable estimates of TAR, TIR, TBR.

\begin{table*}[htbp]
\centering
\small
\setlength{\tabcolsep}{6pt}
\renewcommand{\arraystretch}{1.2}
\begin{tabular}{|l|c|c|c|c|c|}
\hline
\text{Model} & \text{CGM RMSE} & \text{Overall $\text{R}^2$} & \text{TAR MAE} & \text{TIR MAE} & \text{TBR MAE} \\
\hline
SMBG Baseline (No-Interp) & 0.0994 & 0.8638 & 0.0933 & 0.0972 & 0.0220 \\
Full Dual Path (Ours)         & 0.0500 & 0.6912 & 0.0476 & 0.0465 & 0.0163 \\
\hline
\end{tabular}
\caption{Baselines and overall performance on the \textit{repbg} test set.}
\label{tab:baseline_overall}
\end{table*}


\section{Conclusion}


In this study, we proposed DPA-Net, a Dual-Path Spatial–Channel Self-Attention Neural Network, to estimate AGP glycemic metrics directly from SMBG data and thereby provide clinically meaningful insights without requiring CGM. Leveraging dataseta from Jaeb Center Public Data Repository, we addressed the inherent bias and sparsity of SMBG through two strategies. First, we developed an active point selector that identifies the most informative time points for blood glucose measurement, reflecting realistic patient behaviors. Second, we designed a dual-path neural architecture consisting of a reconstruction path and a multi-scale ResNet path for direct metric estimation. By enforcing alignment between the two paths, the model achieves synergy that mitigates bias and improves predictive accuracy.
To the best of our knowledge, this is the first supervised machine learning framework that estimates AGP metrics from SMBG data. Our results demonstrate that DPA-Net achieves high accuracy and unbiased performance, highlighting the potential of SMBG-based AGP estimation as a cost-effective and accessible alternative for diabetes management, particularly in settings where CGM is not widely available.

\appendix
\section*{Appendix A. The network structure for AETCN Active Point Selector}
\label{sec:appendix}
In this appendix, we provide the detailed network architecture of the AETCN Selector. The AETCN Selector consists of $L_3$ stacked temporal convolutional blocks, each containing $B$ branches with different dilation rates to capture glucose fluctuation patterns at varying time scales\cite{bai2018empirical}. A lightweight channel attention module is appended to each block to dynamically reweight the importance of the output feature channels\cite{hu2018squeeze}.

The model input is
\begin{equation}
    \mathbf{X} \in \mathbb{R}^{2 \times T}, \quad T=288,
\end{equation}

where the two channels correspond to the raw CGM values $\mathbf{x}_{\text{CGM}}$ and a positional encoding $\mathbf{p}$.

Each temporal block applies parallel dilated 1D convolutions with dilation rates $\{d_1,\dots,d_{B}\}$. 
Given kernel size $k$ and input $\mathbf{X}$, the $i$-th branch computes
\begin{equation}
\mathbf{Y}^{(i)}_t
= \sum_{j=0}^{k-1} w^{(i)}_j \cdot X_{\,t - d_i \cdot j}.
\end{equation}

Here, $\mathbf{Y}^{(i)}_t$ denotes the output of the $i$-th branch at temporal position $t$, 
$w^{(i)}_j$ is the convolution weight associated with kernel offset $j$, 
$d_i$ is the dilation rate assigned to the $i$-th branch, 
and $X_{\,t - d_i \cdot j}$ represents the input element at time index $t - d_i \cdot j$, 
determined jointly by the dilation rate $d_i$ and kernel offset $j$. 

Branch outputs are concatenated along the channel dimension and fused by a convolution:
\begin{equation}
    \mathbf{Y}_{\text{fused}}
= \mathrm{Conv1D}_{1\times 1}\!\big(\mathrm{Concat}(\{\mathbf{Y}^{(i)}\}_{i=1}^{B})\big).
\end{equation}

A lightweight channel-wise attention then produces a gating mask and reweights features:
\begin{equation}
\mathbf{g} = \sigma\!\big(\mathrm{Conv1D}_{1\times 1}(\mathbf{Y}_{\text{fused}})\big),
\qquad
\mathbf{Y}_{\text{att}} = \mathbf{Y}_{\text{fused}} \odot \mathbf{g},
\end{equation}

where $\sigma$ is the sigmoid function and $\odot$ denotes element-wise multiplication.

A prediction head $f_{\theta}(\cdot)$ maps $\mathbf{Y}_{\text{att}}$ to a sequence of per-step importance scores:
\begin{equation}
s_t = f_{\theta}(\mathbf{Y}_{\text{att}})_t, \quad t=1,\dots,T,
\end{equation}
where $s_t \in [0,1]$ denotes the predicted importance score of the $t$-th time step, reflecting the relative value of observing glucose at that temporal position.

Based on these scores, the selector identifies a subset of observation points by solving a top-$K$ optimization problem with a separation constraint:
\begin{equation}
\mathcal{P}
= \arg\max_{\substack{\mathcal{P}\subseteq\{1,\dots,T\}\\ |\mathcal{P}|=K}}
\;\sum_{t \in \mathcal{P}} s_t
\quad \text{s.t.} \quad
|t - t'| \ge \Delta,\ \forall\, t,t' \in \mathcal{P},\ t \neq t',
\end{equation}
where $\mathcal{P}$ is the selected index set, $K=5$ (reflecting $\approx 2.5\%$ of daily slots in real SMBG data), and $\Delta=12$ enforces at least one hour between any two selected time points (5-minute slots).


The resulting set $\mathcal{P}$ is then used to construct masked inputs and the binary mask for the downstream dual-path model.

\bibliographystyle{IEEEtran}
\bibliography{ref} 
  
\end{document}